\documentclass[10pt,twocolumn,letterpaper]{article}
\pdfoutput=1
\usepackage{iccv}
\usepackage{times}
\usepackage{epsfig}
\usepackage{graphicx}
\usepackage{amsmath}
\usepackage{amssymb}
\usepackage{multirow}
\usepackage{booktabs}

\usepackage[breaklinks=true,bookmarks=false]{hyperref}

\iccvfinalcopy 

\def\iccvPaperID{****} 

\ificcvfinal\fi

\begin{document}

\title{Seven Basic Expression Recognition Using ResNet-18}

\author{Satnam Singh\\
{\tt\small satnam.s.singh@fau.de}

\and
Doris Schicker\\
{\tt\small doris.schicker@fau.de}
}

\maketitle
\ificcvfinal\thispagestyle{empty}\fi

\begin{abstract}
We propose to use a ResNet-18 architecture that was pre-trained on the FER+ dataset for tackling the problem of affective behavior analysis in-the-wild (ABAW) for classification of the seven basic expressions, namely, neutral, anger, disgust, fear, happiness, sadness and surprise. As part of the second workshop and competition on affective behavior analysis in-the-wild (ABAW2), a database consisting of 564 videos with around 2.8M frames is provided along with labels for these seven basic expressions. We resampled the dataset to counter class-imbalances by under-sampling the over-represented classes and over-sampling the under-represented classes  along with class-wise weights. To avoid overfitting we performed data-augmentation and used L2 regularisation. Our classifier reaches an ABAW2 score of 0.4 and therefore exceeds the baseline results provided by the hosts of the competition.
\end{abstract}

\section{Introduction}

Human faces provide a lot of information about a person. Not only can a person be identified based on their facial patterns, faces also mirror people's feelings and emotions and thus play an important role in nonverbal communication. As such recognizing emotions enables us to understand people, their behavior or wishes. Machines and robots that are capable to read humans emotions could interact with them in a human-centered way. Therefore, effective face expression classifiers would allow a new way of human-machine interaction. However it is important that these algorithms work independent of the respective context or the appearance of the person, be it age or ethnicity as examples. Also the algorithms should be able to work in real life environments. 

The problem of affective behavior analysis in-the-wild is tackled by the second workshop and competition on affective behavior analysis in-the-wild (ABAW2) \cite{Kollias.2021} that follows previous competitions \cite{Zafeiriou.2017,Kollias.2019b,Kollias.2020}. Therefore the in-the-wild audiovisual database Aff-Wild2 \cite{Kollias.2019c} was provided that is an extension of the Aff-Wild database \cite{Zafeiriou.2017}. Aff-Wild and Aff-Wild2 databases have been successfully used to train algorithms for face analysis \cite{Kollias.2019,Kollias.2021b,Kollias.2021c}. Aff-Wild2 contains amongst others annotations for all seven basic expressions (i.e. neutral, anger, disgust, fear, happiness, sadness, surprise). In this paper we want to track the second challenge of the ABAW2 competition: to develop an effective classification algorithm that is capable to read the emotion displayed on a face.

We propose an algorithm based on the ResNet \cite{he2016deep} architecture. We used ResNet-18 shared by Meng et al. \cite{meng2019frame} that was pre-trained on the FER+ expression data set \cite{barsoum2016training}.

\section{Methodology}
In this section we describe the used datasets and pre-processing steps. Further we introduce our proposed method.

\subsection{Dataset}
The face expression recognition plus dataset (FER+) was generated by re-labelling the FER dataset \cite{goodfellow2013challenges} by multiple taggers. This provides a better quality ground truth for the training images. Having multiple taggers allowed for the generation of a probability distribution of emotions for each face image \cite{barsoum2016training}.

To train and validate our model we used the database Aff-Wild2 \cite{Kollias.2019c}. This data consists of 564 videos of around 2.8M frames along with the annotations of the seven basic expressions. Out of these, 253 videos are part of the training set and 70 belong to the validation set that were provided by the competition hosts. The remaining 223 videos were part of the test set for which the labels were not disclosed.

\begin{table*}[ht]
    \centering
        \caption{Number of images showing the respective face expressions for the train and validation set provided for the competition (original set) and after under-sampling the dominant classes and over-sampling the under-represented classes (sampled set).}
        \begin{tabular}[t]{*{9}{c}}
            \toprule
            \multicolumn{1}{c}{Data set} &
            \multicolumn{7}{c}{Expression} &
            \multicolumn{1}{c}{Total Number of Images} \\
            \cmidrule(lr){2-8}
             & Neutral & Anger & Disgust & Fear & Happiness & Sadness & Surprise & \\
            \midrule
            Original Train Set & 218364 & 22808 & 12472 & 10847 & 150932 & 101378 & 40353 & 557154     \\
            Sampled Train Set & 43673 & 39914 & 32739 & 32541 & 37733 & 33793 & 40353 & 260746     \\
            \midrule
            Validation Set & 96878 & 8092 & 5401 & 9646 & 54271 & 38983 & 26115 & 239386  \\
            \bottomrule
        \end{tabular}
        \label{Tab:comb}
\end{table*}

\subsection{Pre-processing}
We used the cropped and aligned images that were provided by the ABAW2 competition organizers for training our model.
While pre-processing the dataset, i.e. removing images that corresponded to label -1, it was also found that the number of labels provided did not match the number of images provided. The number of images in the train and validation set for different emotion classes are displayed in Table \ref{Tab:comb}.

To counter class-imbalance between different types of expressions, the dominant classes, namely, Neutral, Happiness and Sadness were under-sampled while the under-represented classes were over-sampled and class-wise weights were added to the cross-entropy loss function (Table \ref{Tab:comb}).
Additionally, the cropped and aligned train set was further divided into a train and validation set in the ratio 80:20. Since the test set labels were not provided, the final trained model was tested on the original validation set provided by the competition hosts to evaluate performance.

\subsection{Model}
Our base model is a ResNet-18 architecture pre-trained on the FER+ expressions data set provided by Meng et al. \cite{meng2019frame}. The hyper-parameters chosen for training on the Aff-Wild2 dataset are summarized in Table \ref{Tab:hyp}. The model was trained on a server with an Nvidia RTX Quadro 8000 GPU with 48GB of memory.

\begin{table}[ht]
    \centering
        \caption{Hyper-parameters chosen for training the Resnet-18 model.}
        \begin{tabular}[t]{lcc}
            \toprule
            Hyper-parameter & Value \\
            \midrule
            {\centering Learning Rate} & $3e^{-3}$      \\
            {\centering Batch Size} & 256  \\
            {\centering Epochs} & 75 \\
            {\centering Rate decay gamma} & 0.1  \\
            {\centering Optimizer weight decay} & $3e^{-4}$ \\
            \bottomrule
        \end{tabular}
        \label{Tab:hyp}
\end{table}

The optimizer chosen was the Adam optimizer  \cite{kingma2014adam} with a weight decay, implemented in PyTorch (https://pytorch.org). Moreover, to avoid over-fitting, data-augmentation was performed by randomly flipping the images with the probability of 0.5 and randomly rotating images up to $\pm{10}$ degrees and finally, the input images were also normalized before being passed into the network. Additionally, a learning rate scheduler was employed to reduce the learning rate after every 15 epochs.

\section{Results}
For evaluating the performance of our method on the provided data set, we calculated the class-wise F1 score and the overall accuracy. The scores required by ABAW2 are calculated as follows:

\begin{equation}
\mathrm{score} = 0.33 * \mathrm{accuracy} + 0.67 * \mathrm{F1}
\end{equation}
Here, accuracy refers to the overall accuracy of the model, i.e., the fraction of images predicted with the correct label to the total number of images in the validation set. The F1 score refers to the macro F1 value, i.e. the unweighted mean of F1 scores for all classes. The F1 score is calculated as:

\begin{equation}
\mathrm{F1} = \frac{2\times Precision \times Recall}{Precision + Recall}
\end{equation}

The results are summarised in Table \ref{Tab:res} and our model exceeds the baseline provided by the organizers for this task.

\begin{table}[ht]
    \centering
        \caption{Results on the validation set.}
        \begin{tabular}[t]{lcc}
            \toprule
            Metric & Value \\
            \midrule
            {\centering Overall Accuracy} & 0.521      \\
            {\centering Macro F1 average} & 0.33  \\
            {\centering Weighted F1 average} & 0.5 \\
            {\centering Score} & 0.4004  \\
            \bottomrule
        \end{tabular}
        \label{Tab:res}
\end{table}

\section{Conclusion}
We contributed to the ABAW2 competition: 7 basic expression classification challenge. Therefore we pre-processed the Aff-Wild2 dataset to counter class-imbalance and over-fitting. We then adapted a ResNet-18 architecture pre-trained on the FER+ dataset for this task. Our method reaches 52.10\% overall accuracy on the validation set as well as a macro F1 average score of 0.33 and thereby a final score of 0.4004.  Our performance thus exceeds the baseline results provided by the organizers in all three metrics.

{
\small
\bibliographystyle{ieee_fullname}
\bibliography{egbib}
}

\end{document}